\documentclass{article}
\usepackage{spconf,amsmath,graphicx}
\usepackage{caption,booktabs}

\usepackage[justification=centering]{caption}
\usepackage[shortlabels]{enumitem}
\setlist[enumerate]{noitemsep}

\usepackage{etoolbox}
\patchcmd{\thebibliography}
  {\settowidth}
  {\setlength{\itemsep}{0pt plus 0.1pt}\settowidth}
  {}{}
\apptocmd{\thebibliography}
  {\small}
  {}{}

\title{LeukoNet: DCT-based CNN architecture for the classification of normal versus Leukemic blasts in B-ALL Cancer}

\name{Simmi Mourya$^{1 \star}$\qquad Sonaal Kant$^{1 \star}$\qquad Pulkit Kumar$^{1 \star}$\thanks{*All authors contributed equally.  $\dagger$ Corresponding Authors} \qquad Anubha Gupta$^{1 \dagger}$ \qquad Ritu Gupta$^{2 \dagger}$}
\address{$^{1}$ SBILab, Department of ECE Indraprastha Institute of Information Technology-Delhi (IIIT-D), India \\
     $^{2}$Laboratory Oncology Unit, Dr. B.R.A.IRCH, AIIMS, New Delhi 110029, India }

\begin{document}
%
\maketitle
\begin{abstract}
Acute lymphoblastic leukemia (ALL) constitutes approximately 25\% of the pediatric cancers. In general, the task of identifying immature leukemic blasts from normal cells under the microscope is challenging because morphologically the images of the two cells appear similar. In this paper, we propose a deep learning framework for classifying immature leukemic blasts and normal cells. The proposed model combines the Discrete Cosine Transform (DCT) domain features extracted via CNN with the Optical Density (OD) space features to build a robust classifier. Elaborate experiments have been conducted to validate the proposed LeukoNet classifier.
\end{abstract}
\section{Introduction}
\label{sec:intro}
\vspace{-0.5em}
Cancer is a deadly disease and is responsible for millions of deaths across the globe every year. Acute Lymphoblastic Leukemia (ALL) constitutes approximately 25\% of the pediatric cancers \cite{bhojwani2015biology}. Timely diagnosis with appropriate treatment can help with increasing the overall survival of the cancer patients. However, high costs of the diagnostic tests pose challenges in providing access to the under-privileged population, particularly, in rural areas. 

In the recent past, convolutional neural networks (CNNs) have achieved immense success in various domains such as speech processing, computer vision, natural language processing, etc. Since CNN architectures have translational invariance and an ability to learn both fine and coarse level features, these architectures are well-suited for such tasks. For example, CNNs have been employed in medical imaging for various tasks such as segmentation \cite{ronneberger2015u}, classification \cite{duggal2017sd} , and object detection \cite{cruz2013deep}. 

Since the existing diagnostics tests based on flow cytometry for the identification/counting of cancer versus normal blasts in cancer are expensive, an automated machine learning based classifier would be cost effective and easily deploy-able in urban as well as rural areas.  
The problem is challenging because 1) the cells are not distinguishable under the microscope, 2) the subject level variability may render the classifier ineffective on prospective subjects' data, and 3) the classification accuracy is desired on both the classes equally. Incorrect diagnosis of either true positive or true negative may prove fatal for a subject, if the chemotherapy is not administered when required or if a subject is administered extra chemotherapy sessions when he/she cannot handle.   

Deep learning based methods can help address all the three above challenges because they extract desirable features from the raw data itself and the architectures are generally scalable. 
In this paper, we propose a novel CNN architecture to classify immature leukemic blasts versus hematogones (normal cells) that would be helpful in the treatment of B-ALL (B-lineage acute lymphoblastic leukemia), a type of white blood cell (WBC) cancer. For the sake of simplicity, we name the immature leukemic blasts as cancer WBCs and hemotagones as normal WBCs throughout the rest of the paper. We use the Stain Deconvolution Layer (SD Layer) as proposed by Duggal et al. \cite{duggal2017sd} and extend the work with the below contributions:
\vspace{-0.5em}
\begin{enumerate}
    \item We add a Discrete Cosine Transform (DCT) layer on the quantity images in the OD space, obtained after the SD-layer. Since flow cytometry detects cells using spectrometry, we believe that there may be useful features in the frequency domain that can be captured by DCT layer-appended CNN architecture.
    \item Further, we use a hybrid framework by combining both DCT layer-appended CNN architecture and without DCT layer-appended CNN architecture, ensuring both frequency domain and spatial domain features are captured appropriately.
    \item We use bilinear pooling instead of average pooling after the last conv layer because it is found to be helpful for fine-grained visual recognition tasks \cite{cruz2013deep}.
\end{enumerate}
\vspace{-1em}
\section{Related Work}
\vspace{-0.5em}
 A lot of research work is done on microscopic image segmentation, detection and classification. Particularly, for the detection of cancer, extensive research has been carried out. Xu et al. \cite{xu2016stacked} used stacked sparse autoencoders for detection of nuclei in breast cancer histopathology images. Zhang et al. \cite{zhang2017deeppap} proposed a deep convolutional network for directly classifying  cervical cells into normal and abnormal cells and directly classified cells  without segmenting them. Cruz-Roa et al. \cite{cruz2013deep} proposed a deep learning architecture through which the technique exploited the features in the intermediate layers as digital staining and used it on on basal-cell carcinoma detection whereas Wieslander et al. \cite{wieslander2017deep} proposed to detect the changes due to malignancy by using Deep Neural Networks. 

In the past, Singhal et al. \cite{singhal2014local} and Zhao et al. \cite{zhao2017automatic} proposed machine learning algorithms using hand crafted features for  Acute Lymphoblastic Leukemia (ALL). Recently, Stain Deconvolution Layer (SD Layer) was proposed in for the classification of cancer versus normal cells. Rather than training in RGB space, classifier was trained on the images in the Optical Density (OD) space. 
This work in \cite{duggal2017sd} is closest to our work because the classifier is designed on the same dataset. However, there are some fundamental differences. While the classifier was built in \cite{duggal2017sd} by mixing cell image data of all subjects, we have separated the cell images at the subject level and the classifier is trained on some subjects and tested on other subjects. This ensures that no cell of a subject in the test data has been shown to the classifier during training. In this manner, we have attempted to learn the subject level variability to ensure reasonably consistent results on the prospective unseen subjects' data. In addition, while 
a dataset of approximately 9000 cell images was used in \cite{duggal2017sd}, we have used a dataset of approximately 14000 cell images. 

\begin{figure}[!ht]
\centerline{
\includegraphics[scale=0.28, trim=0 0 0 10]
{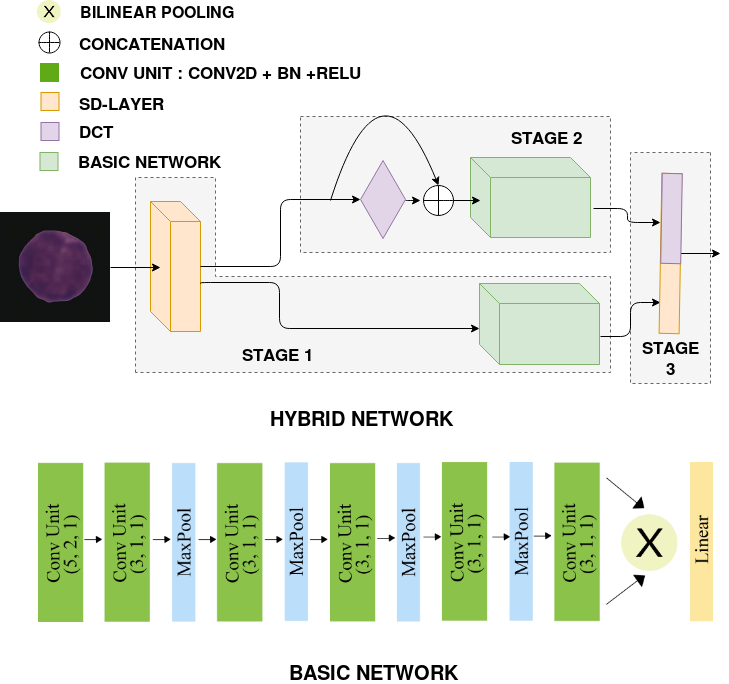}}
\caption{\small Leukonet Architecture}\label{fig:02}
\vspace{-2em}
\end{figure}

\section{Dataset Description}
\vspace{-0.5em}
\label{sec:pagestyle}
Microscopic images were captured from bone marrow aspirate slides of patients diagnosed with B-ALL. 
Images were captured in raw BMP format with a size of 2560x1920 pixels. 
Normal data has been collected from subjects who did not suffer with cancer and hence, the ground truth labels are 100\% correct for this class. Cancer cells have been segmented from the microscopic images collected from the patients who were initially diagnosed with cancer.


We prepared a dataset of 9211 cancer cells from 65 subjects and 4528 normal cells from 52 subjects. 
These images were normalized, the cells of interest were marked with labels by the expert oncologist and segmented. A waiver for written informed consent was obtained from the Ethics Committee of AIIMS, New Delhi, on this dataset for research purposes. 

We split the training data of 13739 cells (9211+4528) into four folds for cross validation such that each fold contains approximately equal proportion of images of both the classes across folds. Also, the folds are prepared at the subject level, i.e., all cells belonging to the same subject are placed in the same fold. In addition, we used test data that is not a part of any fold above and is collected besides the $\approx$14K cells mentioned above, collected from different subjects. Our test data consisted of 312 cancer cells from 2 cancer subjects and 324 normal cells from 2 normal subjects.

Since WBCs were of varied sizes (roughly 250-300 in each dimension), we positioned each cell in the middle of a blank image of size 350x350 with centroid of cell at the centre. We refrained from downsizing the images to avoid any loss of information. During training, we applied augmentations such as rotation, horizontal and vertical flip randomly. We also experimented with augmentations such as shear, Gaussian blur, random crop, and perspective transform. Out of these, first two worked best on our dataset and hence, were finally chosen. We have used shear augmentation within a range of [-20, 20] and Gaussian blur augmentation within a range of [0.0, 0.75]. These augmentation significantly helped in improving overall training and hence, the accuracy. 

\section{Leukonet Architecture}
\label{sec:majhead}
\vspace{-0.5em}
Since limited dataset is available, we trained a small network from scratch instead of working with deep architectures such as Resnet \cite{he2016deep} or DenseNet \cite{huang2017densely} to prevent overfitting. We trained a network with a strided convolutional layer at the front end followed by 5 convolutional layers with batch-norm, maxpool and Relu as shown in Fig 1. As proposed by BCNN \cite{lin2015bilinear}, bi-linear pooling was used at the end of the network where the output of the last convolutional layer were flattened along the spatial dimension and the outer product with its transpose is passed through a linear layer for final  classification. We refer to this network as the Basic Network.

As proposed in \cite{duggal2017sd}, the images were trained in the Optical Density (OD) space rather than RGB space and used the Stain Deconvolution Layer (SD Layer) \cite{duggal2017sd} appended at the beginning of our Basic Network. This combination is further mentioned as Stage-1. 

To emulate the working of flow cytometry, we used Discrete Cosine Transform (DCT) as we believe that there may be useful features in the frequency domain that can be captured by DCT. Two-dimensional DCT was applied on the output of the SD Layer with energy threshold of 95\% reducing the values below the threshold to 1. We further log-normalized the output using log10, while maintaining the sign of the DCT coefficients. This along with the Basic Network is referred to as Stage-2.

Since DCT is a non-linear combination of the pixel values, we believe that the information after the SD layer will be different than that in the DCT.  As an experiment, the three channels of DCT features after Stage-2 were concatenated with the three channel features after Stage-1 of the SD Layer. The concatenated six channel features were fed as input to the Basic Network, which we refer as Stage-2C.

Experiments showed that different cells were detected erroneously from Stage-1 and Stage-2. Hence, to incorporate the knowledge of both these stages, we extracted features from both and trained a single Fully Connected (FC) Layer. This combination ais referred as Stage-3. Further, we experimented with combining Stage-1 and Stage-2C, which we refer to as Stage-3C.  
\vspace{-0.5em}
\section{Experiments}
\vspace{-0.5em}
\label{sec:print}

This work is build upon the work in \cite{duggal2017sd}. At first, the SD Layer of \cite{duggal2017sd} was combined  with the VGG-11 network \cite{simonyan2014very}. We observed overfitting of the model showing an increase in the validation loss during training. We hypothesized that this happened due to a large number of parameters in the network. Furthermore, the network used global average pooling in the last convolutional layer to reduce features. However, this could result in a loss of fine grained information. 

To counter the issue of large number of parameters, we used the basic network shown in Fig.1. This helped us reduce the number of trainable features considerably. Also, to capture the fine grained information, we introduced a binary pooling layer presented in \cite{lin2015bilinear}. We refer to this as `Stage-1 w/o aug' experiment, since we did not introduced any augmentations in the training data yet. This resulted in an increase of 2\% in accuracy on the 3rd validation fold (Table-1), thus validating our hypothesis.  However, we observed that although the network performed really well on the cancer class, the network performed poorly on the normal class. As a result, a lot of fluctuations in the validation loss was also observed. Results of `Stage-1 w/o aug' on all four validation sets and the test set are tabulated in Table 2.

\begin{table}[!ht]
\centering
\caption{\small Experimental results on Val-3.\protect\linebreak $F_{1,N}$: $F_1$ score on Normal class, $F_{1,C}$: $F_1$ score on Cancer class.}
\vspace{-0.5em}
\scalebox{0.99}{
\begin{tabular}{|l|l|l|l|}
\hline
Model Names            & Accuracy   & $F_{1,N}$ & $F_{1,C}$ \\ \hline
\textbf{OD w/ VGG 11}           & 87.07          &  82.76          &   89.66         \\ \hline
\textbf{Stage-1 w/o aug}    & 88.92          & 84.56           &  91.36         \\ \hline
\textbf{Stage-1 w/ aug}    & 90.22        & 85.78           &  92.55          \\ \hline
\textbf{Stage-1 PRelu w/ aug}    & 86.58          & 81.75           &  89.39          \\ \hline
\textbf{Stage-1 PTelu w/ aug}    & 88.17          &  84.09           &  90.59         \\ \hline
\textbf{Stage-1 w/ Normal aug}     &  89.18         &  85.01         &  91.54           \\ \hline
\textbf{Stage-2C w/o aug}        &  87.56          &   82.39         &   90.38        \\ \hline
\textbf{Stage-2C PRelu w/o aug} &  84.28         &   78.79         &  87.51          \\ \hline
\textbf{Stage-2C PTelu w/o aug} &  86.39         &    81.32       &  89.29           \\ \hline
\textbf{Stage-3C w/ Normal aug}    & \textbf{89.70} &  \textbf{85.84} & \textbf{91.95}   \\ \hline
\textbf{Stage-3 w/ Normal aug}     &   87.20          &   82.20      &   90.01            \\ \hline
\end{tabular}}
\small{Note: ReLU activation function is the default if not mentioned with the model.}
\vspace{-1em}
\end{table}
\begin{table*}[!ht]
\caption{\small Experimental results on all four folds of validation and the test set. \protect\linebreak $F_{1,N}$: $F_1$ score on the Normal class, $F_{1,C}$: $F_1$ score on the Cancer class.}
\vspace{-0.5em}
\scalebox{0.83}{
\begin{tabular}{|l|l|l|l|l|l|l|l|l|l|l|l|l|l|l|l|}
\hline
\textbf{Model name}      & \multicolumn{3}{c|}{\textbf{Val0}}                 & \multicolumn{3}{c|}{\textbf{Val1}}                 & \multicolumn{3}{c|}{\textbf{Val2}}                 & \multicolumn{3}{c|}{\textbf{Val3}}                 & \multicolumn{3}{c|}{\textbf{Test set}}             \\ \hline
                         & Acc   & $F_{1,N}$ & $F_{1,C}$ & Acc   & $F_{1,N}$ & $F_{1,C}$ & Acc   & $F_{1,N}$ & $F_{1,C}$ & Acc   & $F_{1,N}$ & $F_{1,C}$ & Acc   & $F_{1,N}$ & $F_{1,C}$ \\ \hline
\textbf{Stage-1 w/o aug}  & 93.00             & 88.57           & 94.95           & \textbf{93.03} & 84.09           & 93.03           & 92.63          & 87.06           & 94.84           & 88.92          & 84.56           & 91.36           & 88.08          & 87.94            & 88.2           \\ \hline
\textbf{Stage-1 w/ aug}   & 93.93          & 90.58           & 95.52           & 91.71          & 86.41           & \textbf{94.03}  & 93.3           & 88.14           & 95.33           & \textbf{90.22} & \textbf{85.78}  & \textbf{92.55}  & 88.54          & 88.88           & 88.17           \\ \hline
\textbf{Stage-2 w/o aug}  & 92.37          & 87.62           & 94.49           & 88.97          & 82.13           & 92.02           & 88.24          & 79.87           & 91.69           & 85.71          & 79.19           & 89.11           & 81.11          & 80.51           & 81.68           \\ \hline
\textbf{Stage-2C w/o aug} & 93.28          & 89.58           & 95.04           & 91.57          & 86.97           & 93.77           & 88.69          & 80.63           & 91.58           & 87.56          & 82.39           & 90.38           & 83.74          & 84.06           & 83.41          \\ \hline
\textbf{Stage-3 w/o aug}  & \textbf{94.95} & \textbf{91.95}  & \textbf{96.32}  & 88.94          & 82.49           & 91.92           & 92.43          & 87.21           & 94.62           & 86.26          & 79.81           & 89.58           & 87.77          & 88.08           & 87.44          \\ \hline
\textbf{Stage-3  w/ aug}  & 94.39          & 91.38           & 95.84           & 90.28          & 84.45           & 92.93           & \textbf{94.34} & \textbf{90.47}  & \textbf{95.98}  & 87.00             & 81.50            & 89.98           & 89.47          &    89.88        &  89.03        \\ \hline
\textbf{Stage-3C w/o aug} & 94.22          & 91.02           & 95.72           & 91.57          & \textbf{87.24}  & 93.70            & 91.89          & 86.18           & 94.27           & 88.37          & 83.52           & 91.01           & \textbf{89.62} & 89.89  &  \textbf{89.34}          \\ \hline
\textbf{Stage-3C w/ aug}  & 94.05          & 90.99           & 95.55           & 91.51          & 87.19           & 93.65           & 93.98          & 89.79           & 95.73           & 88.82          & 84.10            & 91.38           & \textbf{89.62} &   \textbf{90.24}   & 88.92  \\ \hline
\end{tabular}}
\end{table*}

Our dataset is skewed in class ratio with the ratio of cancer WBCs:Normal WBCs of about 2:1.  We believe this to be the main reason behind the low performance on the Normal class and the observed fluctuations in the validation loss. To resolve this, we introduced augmentations in the data that, in our opinion, simulate conditions of real experiment data preparation. It was observed that affine augmentation like shear applied in a range of [-20, 20] works best for our dataset. This is possibly because cells do get sheared upto some extent, while the microscopic slides are prepared. Also, there can be variations in the sharpness of the image because manual focusing of microscope lens was carried out during image capture. Thus, we used Gaussian blur in the images to induce the effect. We used the augmented version of Stage-1 model for this experiment, referred as, `Stage1 w/ aug' throughout this paper. 

Since the number of normal WBCs were less, we also applied the above mentioned augmentations to only Normal class and then randomly sampled the images choosing the number of images equal to those of Cancer class. We call this experiment as `Stage-1 w/ Normal aug'. This helped in further reduction of the fluctuations, indicating better training. Although fluctuations on the validation set reduced to some extent, they were still observed. 

Furthermore, to find the reason behind the fluctuations on the validation set, we analyzed the mistakes made by the `Stage1 w/ aug' model. By analyzing the results carefully, we noted that our model was failing at the subject-level data, i.e., either most of the cells of a subject were recognized correctly or were mis-classified. This shows that there is subject variability in the data and hence, needed to be addressed carefully. On the contrary the standard lab test of Flow Cytometry provides consistent results. 

To emulate the Flow Cytometry test that works on the principle of flow cytometry, we took the Discrete Cosine Transform (DCT) on the OD space values of the  images and trained the CNN model on the DCT coefficient images. We call this experiment, `Stage-2 w/o aug' since it combines Stage-1 architecture with DCT enhancement. Although we observed a decline in the accuracy compared to `Stage-1 w/o aug' (Table-2), some of the hard examples not classified correctly by `Stage-1 w/o aug' were now being correctly classified by the `Stage-2 w/o aug' model. Moreover, the fluctuations observed earlier were reduced considerably.  
Since DCT is a non-linear combination, we believe that the data before and after DCT operation may carry different information. Thus, the features learned (3 image planes) on DCT-appended CNN were concatenated with the features learned by the CNN on the direct OD space (3 image planes). The concatenated six channel features were used as input to our `Stage-2C w/o aug' model and compared to `Stage-2 w/o aug' model. This shows an increase in accuracy (Table-2) with reduction in fluctuations. 


We also hypothesized that the unstable training could also be a result of incorrect activation function. Therefore, to validate this hypothesis, we ran `Stage-1 w/aug' with  P-TELU activation as proposed by Gupta et al. \cite{gupta2017p} and PReLU activation function as proposed by He et al. \cite{he2015delving}. Although this increased the parameters marginally, this setup could help us learn more significant features. In Table 1, we report the results by using these two activation functions on previous best models, namely, `Stage-1 w/aug' and `Stage-2C w/o aug'. From the results of Table 1, it can be concluded that for our problem, ReLU performed best compared to PReLU and P-TELU.

We further analyzed the results of `Stage-2C w/o aug' model and `Stage-1 w/ aug' on validation set-3. Interestingly, we observed that the sets of samples on which both the models were failing, were almost mutually exclusive. Thus, we combined the two aforesaid models and developed a new hybrid model. We call this hybrid model, `Stage-3C w/o aug',  in which we concatenate features from both the networks and then train a single fully connected (FC) layer. This experiment relied on our hypothesis that, if combined, these two models would collectively perform better as compared to the individual models. To further validate our hypothesis and obtain thorough results, we replicated this configuration for the other possible hybrid of set (`Stage-2 w/o aug', `Stage-C w/o aug') and set (`Stage-1 w/o aug', `Stage-1 w/aug'). Models resulting from the hybrid of `Stage-1 w/o aug' with `Stage-2 w/o aug' and `Stage-1 w/o aug' with `Stage-2C w/o aug' are denoted as \textbf{`Stage-3 w/o aug'} and \textbf{`Stage-3C w/o aug'}, respectively. The `w/o' tag in the hybrid experiment represents the training configuration of corresponding Stage-1 model. Since we did not achieve good results with `Stage-2C w/ aug' experiment, we did not proceed with augmentation on Stage-2. Similarly, hybrids obtained from  `Stage-1 w/ aug'  with `Stage-2 w/o aug' and `Stage-1 w/ aug' with `Stage-2C w/o aug' are denoted as \textbf{`Stage-3 w/ aug'} and \textbf{`Stage-3C w/ aug'} respectively. We report the results of all the mentioned hybrids in Table 2. Finally, from Table 2, we observe that \textbf{`Stage-3C w/ aug'} configuration performs best on our test set. Furthermore, we experimented with only Normal cell augmentation for some of our hybrid models, namely `Stage 3 w/ Normal aug' and `Stage-3C w/ Normal aug'. Since results of these hybrids on validation set 3, as referred from Table 1, were not better than the aforementioned hybrids, we did not continue with further experimentation on these hybrids. 

Each training experiment took approx. 7-10 hours on an NVIDIA GeForce GTX 1080 Ti GPU, depending upon the amount of augmentation and depth of the network architecture. We terminated the training process when the accuracy on the validation set was saturated. Models resulting from best validation set accuracy were used on the test set.   
\vspace{-0.5em}
\section{Conclusion}
\vspace{-0.5em}
In this work, we present a novel CNN architecture, namely, Leukonet  for the classification of cancer cells from normal cells in white blood cancer. We employed a hybrid architecture that utilizes the knowledge of two differently trained models. This improves the classification accuracy. To achieve this, we fuse the features of CNN model trained on the DCT of the OD space images with the features of CNN model trained on the OD space. Model were trained with and without data augmentation separately and tested exhaustively. Augmentations of shear and blur present in the real data guided us to choose these augmentation techniques.  Finally, the best of the models were used to train the hybrid architecture ensuring that individually learned representations compliment each other. In the future work, we will try to improve our results by generating more training images using Generative adverserial network (GAN) that can provide better learning and help in overcoming the problem of small medical imaging data.
\vspace{-0.5em}
\section{Acknowledgements}
\vspace{-0.5em}
Authors gratefully acknowledge the research funding support (Grant: EMR/2016/006183) from the Department of Science and Technology, Govt. of India for this research work. The funder had no role in study design, data collection and analysis, decision to publish, or preparation of the manuscript.



\bibliographystyle{IEEEbib}
\bibliography{strings,refs}

\end{document}